\documentclass[sn-aps]{sn-jnl}

\usepackage{graphicx}%
\usepackage{multirow}%
\usepackage{amsmath,amssymb,amsfonts}%
\usepackage{amsthm}%
\usepackage{mathrsfs}%
\usepackage[title]{appendix}%
\usepackage{xcolor}%
\usepackage{stackengine}
\usepackage{textcomp}%
\usepackage{manyfoot}%
\usepackage{booktabs}%
\usepackage{algorithm}%
\usepackage{algorithmicx}%
\def\delequal{\mathrel{\ensurestackMath{\stackon[1pt]{=}{\scriptstyle\Delta}}}}
\usepackage{algpseudocode}%
\usepackage{listings}%

\theoremstyle{thmstyleone}%
\theoremstyle{thmstyletwo}%

\theoremstyle{thmstylethree}%

\begin{document}

\title[Article Title]{FISTNet: FusIon of STyle-path generative Networks for Facial Style Transfer}


\author*[1]{\fnm{Sunder Ali} \sur{Khowaja}}\email{sandar.ali@usindh.edu.pk}

\author[2]{\fnm{Lewis} \sur{Nkenyereye}}\email{nkenyele@sejong.ac.kr}
\author[3]{\fnm{Ghulam} \sur{Mujtaba}}\email{gmujtaba@ieee.org}


\author[3,4]{\fnm{Ik Hyun} \sur{Lee}}\email{ihlee@tukorea.ac.kr}

\author[5]{\fnm{Giancarlo} \sur{Fortino}}\email{giancarlo.fortino@unical.it}
\author[6]{\fnm{Kapal} \sur{Dev}}\email{kapal.dev@ieee.org}

\affil*[1]{\orgdiv{Faculty of Engineering and Technology}, \orgname{University of Sindh}, \orgaddress{\city{Jamshoro}, \postcode{76080}, \state{Sindh}, \country{Pakistan}}}

\affil[2]{\orgdiv{Department of Computer Science}, \orgname{and Information Security}, \orgaddress{\city{Seoul}, \postcode{05006}, \country{South Korea}}}

\affil[3]{\orgdiv{Department of Computer Science}, \orgname{West Virginia University}, \orgaddress{\city{Morgantown}, \postcode{26506}, \state{West Virginia}, \country{United States of America}}}

\affil[3]{\orgname{IKLab Inc.}, \orgaddress{\city{Siheung-Si}, \postcode{15073}, \state{Gyeonggi-do}, \country{Republic of Korea}}}

\affil[4]{\orgdiv{Department of Mechatronics}, \orgname{Tech University of Korea}, \orgaddress{\city{Siheung-Si}, \postcode{15073}, \state{Gyeonggi-do}, \country{Republic of Korea}}}

\affil[5]{\orgdiv{Department of Informatics, Modeling, Electronics and Systems}, \orgname{University of Calabria}, \orgaddress{\city{Calabria}, \country{Italy}}}

\affil[6]{\orgdiv{Department of Computer Science and ADAPT Centre}, \orgname{Munster Technological University}, \orgaddress{\city{Cork}, \country{Ireland}}}


\abstract{With the surge in emerging technologies such as Metaverse, spatial computing, and generative AI, the application of facial style transfer has gained a lot of interest from researchers as well as startups enthusiasts alike. StyleGAN methods have paved the way for transfer-learning strategies that could reduce the dependency on the huge volume of data that is available for the training process. However, StyleGAN methods have the tendency of overfitting that results in the introduction of artifacts in the facial images. Studies, such as DualStyleGAN, proposed the use of multipath networks but they require the networks to be trained for a specific style rather than generating a fusion of facial styles at once. In this paper, we propose a FusIon of STyles (FIST) network for facial images that leverages pre-trained multipath style transfer networks to eliminate the problem associated with lack of huge data volume in the training phase along with the fusion of multiple styles at the output. We leverage pre-trained styleGAN networks with an external style pass that use residual modulation block instead of a transform coding block. The method also preserves facial structure, identity, and details via the gated mapping unit introduced in this study. The aforementioned components enable us to train the network with very limited amount of data while generating high-quality stylized images. Our training process adapts curriculum learning strategy to perform efficient, flexible style and model fusion in the generative space. We perform extensive experiments to show the superiority of FISTNet in comparison to existing state-of-the-art methods.}




\maketitle

\section{Introduction}\label{sec1}

Facial style transfer is an art form that is extensively used in a wide variety of applications including social media filters, virtual character creation, animation production, advertising, non-fungible tokens (NFTs), and the Metaverse \cite{AI6G}. Many artists in the past have endeavored to create exaggerated or simplified replicas of real-world figures, either in cartoon, anime, or arcane styles. Usually, it takes intensive professional skills and laborious efforts to recreate real-world people in the aforementioned styles for artists. The advent of generative adversarial networks (GANs) has caused a paradigm shift in skill acquisition and artistic media creation through facial style transmission. 

From a macro perspective, the field of facial style transfer belongs to the category of image-to-image translation. GANs have been extensively used for such translation tasks, especially for image-to-anime or other style conversion \cite{AnimeGAN}. Generally, facial style transfer techniques can be categorized into portrait \cite{WarpGAN, MangaGAN, Yi2020} and scene \cite{AnimeGAN,CartoonGAN,Whitebox}-based style transfer methods, which have been applied and realized in various applications. The former is best suited for producing caricature \cite{WarpGAN, StyleCariGAN} and manga \cite{MangaGAN,Yi2020} inspired facial style images. The latter learns a transformation from photo to cartoon or any other style by leveraging pre-extracted representations and specialized losses. Both techniques rely heavily on induced facial landmarks and decomposed facial components, making the aforementioned methods unsuitable for applications centered around common scenes.

Another category for facial style transfer is the StyleGAN \cite{StyleGAN, Usercontrol} or unsupervised image-to-image translation \cite{ugatit}, which addresses the challenging task of selfie2anime. These methods are required to train for each specific style in order to generate a satisfactory output. Many existing studies only considered using single-style facial transfer and generating a high-quality image. However, when switching to multi-style transfer the methods mostly generate unwanted artifacts with acceptable-quality images. A similar problem is also faced by studies that aim to apply diverse styles, such as combining cartoon, anime, and arcane styles, accordingly \cite{Multistylecartoon}. Another problem while performing multiple styles is that they are unable to preserve facial poses, characteristics, and features, accordingly. Misalignment between the facial features of target and source domains causes unwanted artifacts in the facial style transferred output images.

\begin{figure*}[t]
  \centering
   \includegraphics[width=\linewidth]{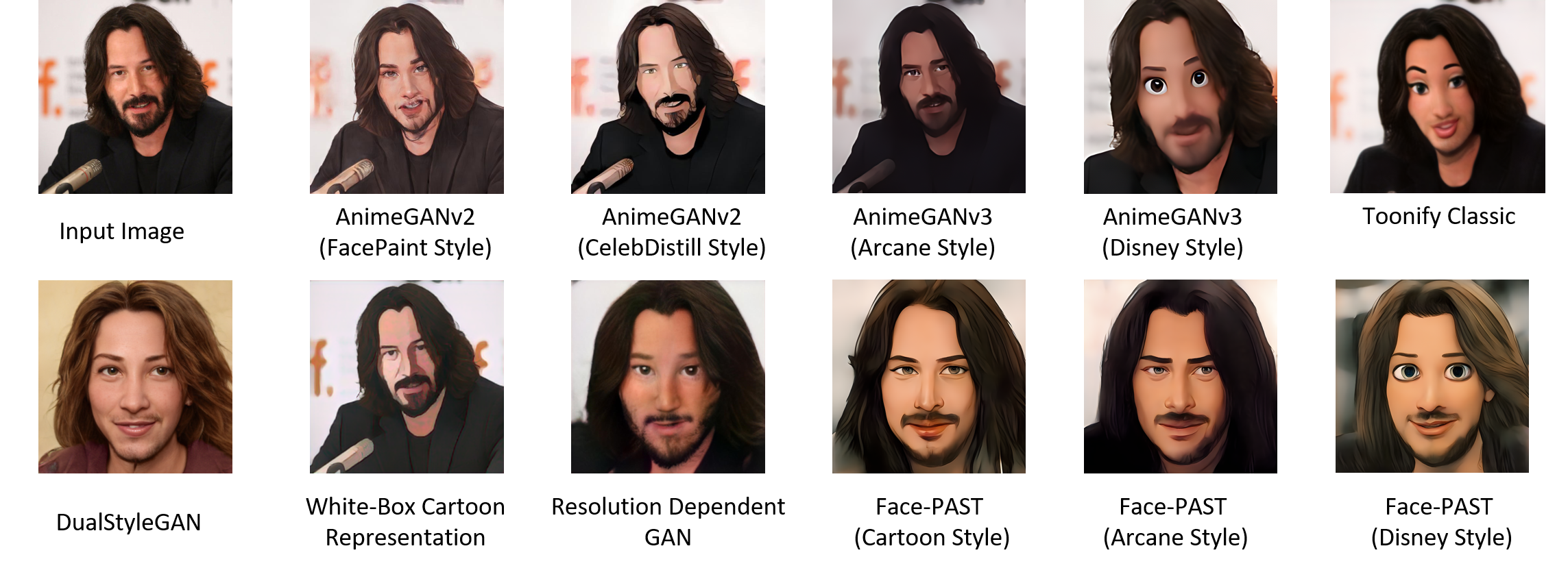}
   \caption{The work proposes FISTNet that performs high resolution facial style transfer, i.e. 1024$\times$1024. The existing works either overfit styles onto the faces that do not preserve facial structure and characteristics, such as Resolution Dependent GAN, DualStyleGAN, and Toonify \cite{VToonify}, or do not transfer diverse styles such as AnimeGAN and white-box cartoon representations \cite{Whitebox}. In addition, studies such as AnimeGAN introduce artifacts into facial images. Aforementioned works can also generate different styles of images along with the proposed work, respectively.}
   \label{fig:fig1}
\end{figure*}

Figure \ref{fig:fig1} shows a qualitative comparison between some existing works and the proposed FISTNet. Some of the works completely transform the facial characteristics while compromising the original facial structure. Meanwhile, others retain the facial structure to a certain extent but fail to propagate high-quality and diversified styles. Furthermore, existing works train the network for each style separately to avoid incorporating multiple styles into the image. One of the seminal works is the multi-style cartoonization \cite{Multistylecartoon} that used multiple encoders for different style networks and multiple discriminators for identifying the cartoon style and selecting the appropriate style loss. Another work is the DualStyleGAN \cite{DualStyleGAN} proposed the use of an external style pass and a progressive fine-tuning strategy to generate high-quality stylized images.

In this regard, we propose a fusion of style-path generative networks (FISTNet) for facial style transfer, a multimodal generative network to introduce an effective multi-style transfer while realizing the control of facial characteristics in images. Inspired by the study DualStyleGAN and multi-style cartoonization, we leverage the hierarchical architecture from StyleGAN to embed multiple styles in fine and coarse-resolution layers via an extrinsic style path. This is done by leveraging the fusion of pre-trained styleGAN networks for embedding diverse styles. The intrinsic style path uses a base style (cartoonization) and residual blocks that retains the facial characteristics. In contrast to the DualStyleGAN, we use the pretrained encoders from existing state-of-the-art stylization networks to produce high-quality images along with gated mapping units to extract the domain-specific features. At times the extrinsic style path alters the behavior of pre-trained networks, which might affect the style transfer as well as the quality of the output image. We show that the fusion of pre-trained models not only helps in the non-alteration behavior of the extrinsic style path but also positively affects the fine-tuning of convolutional layers. Furthermore, we maintain the facial details by considering identity, segmentation, and structural losses in the intrinsic style path. The main contributions of this work are as follows:

\begin{itemize}
\item{We propose FISTNet for facial style transfer while preserving facial structure details.}
\item{We adapt hierarchical style training such as extrinsic and intrinsic style transfer with pre-trained encoders to generate better results.}
\item{We perform the fusion of style networks to generate diverse yet facial preservation style transfer.}
\item{Extensive quantitative and qualitative comparison has been carried out with current state-of-the-art methods.}
\end{itemize}

The rest of the paper is structured as follows: Section \ref{sec:related_work} provides a consolidated review for image stylization in the context of proposed study. Section 3 provides working methodology of the proposed work. Section 4 present experimental results to prove the effectiveness of the proposed approach. Section 5 concludes the work while highlighting future directions.

\section{Related Works}\label{sec:related_work}
This section consolidates a brief review of existing works concerning style transfer with GANs, StyleGAN, and image-to-image translation techniques.
\subsection{Non-Photorealistic Rendering}
The NPR algorithms mainly render 3D shapes to create a cartoon-like effect through cel-shading techniques, however adding cartoon style to existing photos is a much more challenging task \cite{NR1}. The rendering of 3D shapes to impersonate cartoon-like effect leverage optimization and filtering methods. However, by doing so, a high-level abstraction that exhibits an artistic effect is not achieved. Some studies used supplementary segmentation or add user's interaction to generate  cartoon portraits, respectively \cite{NR2}.

\subsection{Style Transfer Networks}
The use of style transfer was introduced to overcome the issues associated with non-photorealistic rendering. The traditional style transfer was proposed to generate stylized images using image training pairs. The use of such methods was able to create good results but was limited to a specific style. Researchers then proposed the use of the VGG network \cite{VGG} for style transfer, which was considered to be good for extracting semantic features. The use of the VGG network also eliminated the issue of using training pairs. The use of neural style transfer was a good idea to start \cite{NST}, but, when using multiple faces, the style is transferred in a homogeneous manner that fails to produce smooth shading or clear edges. The study \cite{Chuan2016} used convolutional neural network (CNN) based feature maps and Markov Random Field for local matching to transfer the style. However, the transference yields semantically incorrect output. A deep analogy-based method was proposed in \cite{Jing2017} to produce semantically correct output while extracting meaningful correspondences. But it was limited to a single style only and at times yielded ambiguity. \\
A dedicated CNN was recently used in the study [Chen ref] to classify between the comic and non-comic images to cope with the aforementioned region ambiguity, but it was still limited to a single style. An alternate and popular approach for style transfer has been generative adversarial networks (GANs) \cite{GAN}. In recent times, several studies have proposed the use of GANs to solve problems concerning pixel-to-pixel image synthesis. At the start, the GANs were considered to be impractical due to the need for a large set of paired images \cite{Isola}. This fundamental issue was addressed by the study CycleGAN \cite{CycleGan} which proposed a way for image translation while using unpaired images. 
The works such as GDWCT \cite{GDWCT} and UNIT \cite{UNIT} perform similarly to CycleGAN while considering special characteristics like clear edges and high-level abstraction. 
One of the seminal works on facial style transfer was the cartoonGAN \cite{CartoonGAN} which proposed the use of content loss to preserve facial characteristics while performing the style transfer but again was limited to a single style. 

\subsection{Image-to-Image Translation}
Image-to-image translation is another technique to perform facial style transfer \cite{SPatchGAN, Advloss}. Such a method relies on the bi-directional mapping between the style domain and the facial image \cite{CycleGan}. Some studies use attention mechanisms to find appearance discrepancies between key regions \cite{ugatit}, while others use shared discriminator layers to extract features that are common in both domains \cite{AniGAN}. The study \cite{GNR} learns style and content image features to generate caricature-style images through facial deformation and image warping, respectively. The problem with bi-directional mapping is that it requires a long training time and generates low-resolution images \cite{DualStyleGAN}. Olivier et al. \cite{newref2} focused on implementing an image-to-image translation method that can adopt 3D facial geometry to neutralize expressive faces. Subsequently, the method uses SpiralNet++ and FUNIT to deform and blendshapes on facial images for style transfer. They referred to their method as FaceTuneGAN. Chen et al. \cite{newref4} extended the work of AnimeGAN \cite{AnimeGAN} and proposed AnimationGAN thet uses bottlenecks in the residual network and hybrid attention mechanism to improve upon the toonification of the facial images, however, its still limited to the style transfer of single style in an implicit manner. Liu et al. \cite{newref8} and Melnik et al. \cite{newref9} conducting an extensive survey that highlights the studies performed for editing and generating faces using StyleGAN and manipualtion of facial attributes. Both of the studies extensively layout the tasks being performed by existing works to manipulate facial attributes using image-to-image translation methods. 

\subsection{StyleGAN}
Recently, StyleGAN was proposed that was able to generate high-resolution images while performing facial style transfer \cite{StyleGAN}. Since the inception of StyleGAN, several studies have considered fine-tuning it to generate plausible results with limited data. The study \cite{RDGAN} used fine-tuned StyleGAN to generate cartoon faces by extracting latent embedding and fine-tuned the model to perform semantic alignment for toonifying the facial image. Some works extended the toonifying approach by training the model on extremely limited data \cite{FewShot}, efficient use of latent code \cite{AgileGAN}, and embedding acceleration \cite{EnStyle}. The study \cite{StyleGAN2} introduced the use of exemplar style images instead of features extracted from fine-resolution-layer to generate better facial style transfer images. However, without valid supervision, the model alignment gets weakened, resulting in an efficient color transfer but does not perform well while preserving structural information. Some studies have proposed the use of cascaded StyleGANs, i.e. using the first StyleGAN to extract style codes and the second to generate the stylized image. These studies produce fixed-size images that hinder the generation of dynamic faces \cite{VToonify}. The StyleGAN3 \cite{StyleGAN3} was proposed to solve the problem of unaligned faces, but it still is limited to the fixed-size image. Recently, a study \cite{DualStyleGAN} proposed DualStyleGAN that uses an extrinsic style path to overcome the model alignment problem and prior facial destylization. Our work is different from the aforementioned ones, as first, it uses pre-trained encoders to generate base and secondary style images, and second, aims to preserve the facial structures and details. A comparison of the results generated using the proposed work with the existing ones is shown in Figure \ref{fig:fig1}. 
  
\begin{figure*}[t]
  \centering
   \includegraphics[width=\linewidth]{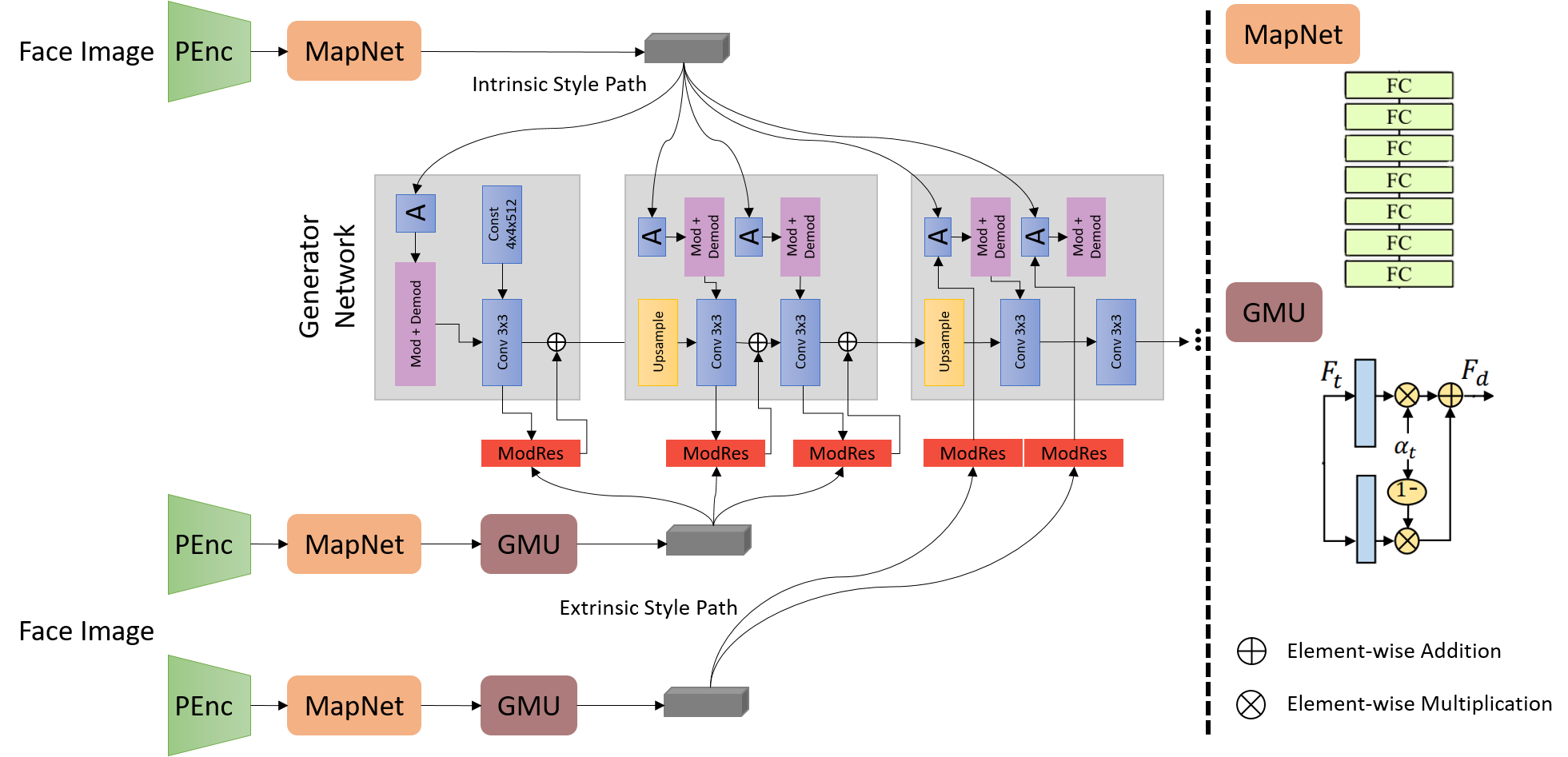}
   \caption{The proposed FISTNet network architecture.}
   \label{fig:network}
\end{figure*}

\subsection{GAN Inversion framework}
Recently, with the rise of generative AI techniques, many works have proposed the manipulation of facial attributes in order to embedded style transfer through GAN inversion frameworks. Liu et al. \cite{newref1} proposed the use of face swapping and regional GAN inversion to perform style fusion. The method leverages the use of styleGAN but does not intrinsically applied style transfer to the base facial image, rather focuses on the swapping of any given image to the reference image. Lan et al. \cite{newref3} proposed a self-supervised learning based strategy to reconstruct 3D shapes and textures from a single 2D image. After a faithful 3D reconstruction of 2D shape, facial style transfer is proposed through global latent codes. The style transfer is not perform to add effects in this study but rather to manipulate the facial attributes. Zheng et al. \cite{newref5} proposed a new joint loss function for a generative adversarial based network that employs cross-fusion attention for style transfer. Their proposed network employed frequency domain loss to incorporate the contextual information between high-level and low-level features for reconstructing the image. Subsequently, the method will use the features to manipulate the image characteristics for the style transfer. Their proposed method works well but does not preserve the facial features as proposed in this method. Ren et al. \cite{newref6} also proposed a cross-modal decoding framework by leveraging styleGAN to reconstruct faces from fMRI data. The method uses multi-level visual information from brain singles and extracts high-level features to be processed by transformation blocks and multi-stage refinement method in order to construct faces. Although the styleGAN have been employed in this study but more focus is redirected towards the use of fMRI data for reconstructing faces rather than performing facial style transfer. Peng et al. \cite{newref7} proposed a dual branch style encoder, namely, interpretable semantic generative adversarial networks (ISFB-GAN) that use inversion techniques to reconstruct the facial geometry for manipulating appearance styles. The focus of this work is to preserve the background information and beautify the facial characteristics like skin and face geometry, rather than adding new styles to the faces.

\section{FISTNet}
The goal of this work is to propose a facial style transfer that adopts pre-trained networks, i.e. StyleGAN \cite{StyleGAN} and AnimeGAN \cite{AnimeGAN}, which would allow the users to characterize multiple styles using a single image while considering a few samples of data. The facial alignment problem is dealt with in the gated mapping unit module while modeling multiple styles problem has been addressed using individual style paths, explicitly. The proposed network is trained using a curriculum learning strategy to perform stable conditional fine-tuning. The details for each block are presented in the subsequent subsections.

\subsection{Intrinsic Style Path}
We use the transfer learning approach to train the Transformer-based StyleGAN model for our intrinsic style transfer path as shown in Figure \ref{fig:network}. The reasons for opting transfer learning approach are twofold: the first is to retain the facial structures and details, and the second is to work with limited data for training. Therefore, performing the fusion of pre-trained StyleGAN and AnimeGAN for fine-tuning is an efficient approach to generate high-quality images. We adopt StyleGAN trained on the FFHQ dataset \cite{StyleGAN} as the pre-trained encoder ($PEnc_{SG}$). Fine-tuning is performed using structural loss between the pre-trained and transferred models. The formulated loss function is shown in equation 1. 

\begin{equation}
\mathit{loss}_{suc} = \frac{1}{K} \sum_{k=1}^{K} \|\mathbb{G}^k_{base}(I) - \mathbb{G}^k_{tl}(I)\|^2
\end{equation}
where $k$ is the index of the StyleGAN block and $\mathbb{G}$ represent the base and transfer generator model, respectively. We also use the adversarial loss for fine-tuning the StyleGAN which is formulated in equation 2. 

\begin{equation}
\mathit{loss}_{adv} = \mathbb{E}_{I \sim p_I}[log(1 - \mathbb{D}(\mathbb{G}_{tl}(I)))] + \mathbb{E}_{i \sim p_{data}}[log(\mathbb{D}(i)]
\end{equation}

The fine-tuning of the StyleGAN module is performed by combining both of the aforementioned losses, i.e. $\mathit{loss}_{isp} = \mathit{loss}_{suc} + \mathit{loss}_{adv}$. However, even with fine-tuning some artifacts are introduced that can affect the facial structure and expressions. In this regard, it is necessary to extract semantic features from the latent space of StyleGAN. The study \cite{Semantics} found the semantic features using a  closed-form factorization, such as $\mathbb{G}(I) \delequal \mathfrak{b}$ + $\mathfrak{w}I = \mathfrak{z}$, where $\mathfrak{b}$ and $\mathfrak{w}$ are biases and weights, respectively. The same study suggested that the latent codes can be manipulated by using the formulation $\mathbb{G}(I+\sigma y) = \mathfrak{z} + \sigma \mathfrak{w} y$, where manipulation intensity is represented by the notation $\sigma$ and semantic attribute representation in latent space is represented by $y$. It was discussed in the same study that the weights in mapping network contains information related to image variations, therefore, the discovery of semantic attributes can be performed by decomposing the weights, accordingly. The decomposition can be performed using the optimization function shown in equation 3.

\begin{equation}
y^* = \operatorname*{argmax}_{y \in \mathbb{R}^d: y^Ty = 1} \|\mathfrak{w}y\|_2^2
\end{equation}

The function $\|.\|_2$ represents the $L_2$ norm. Another problem that occurs during the manipulation of latent code through the mapping network is the change in facial identity, which might hinder the motivation of this work. Therefore, we use an identity loss that optimizes the offset, i.e. $\sigma y^*$ along with a pre-trained recognition network \cite{FaceRec} that is responsible for facial identity regularization. The formulation for the aforementioned optimization problem is shown in equation 4. 

\begin{equation}
\mathit{loss}_{id} = \| f(\mathbb{G}_{base}(\sigma y)) - f(\mathbb{G}_{base}(\sigma y + \sigma y^*)) \|^2
\end{equation}

There might be a condition that yields $\sigma y^* \rightarrow 0$. Such condition may result in insignificant optimization. To overcome the said issue, we also add a constraint that aims to restore the low-level features between the manipulated and input modality. In order to do so, we use segmentation methods \cite{FaceSeg} to extract the facial area, represented by $f_{seg}$. The constraint is defined as a regularization function in equation 5. 

\begin{equation}
\mathit{loss}_{seg} = \| f_{seg}(\mathbb{G}_{base}(\sigma y)) - f_{seg}(\mathbb{G}_{base}(\sigma y + \sigma y^*)) \|^2
\end{equation}
The combined optimization function can then be given as [$\mathit{loss}_{mapnet} = \mathit{loss}_{id} - \alpha_{seg} \mathit{loss}_{seg}$], where $\alpha_{seg}$ represents the hyperparameter to control the trade-off, accordingly.

\subsection{Extrinsic Style Path}
Our external style path simply performs the fusion of two pre-trained networks, i.e. AnimeGANv2 trained on portrait and celeb distill styles to add semantic cues such as facial color, eye styles, hair color, and shapes, accordingly. Different styles including arcane, Disney, and others can be leveraged by performing the fusion of pre-trained networks to produce diverse results as shown in Figure \ref{fig:fig1}. We use the same optimization function for MapNet as discussed in the prior section to retain facial details and structure, respectively. The features extracted using MapNet will then undergo the Gated Mapping Unit (GMU) adopted from the study \cite{GMU}. However, unlike the study, we only use domain-specific features instead of combining them with group-specific features. Let's denote the feature extracted from MapNet as $F_t$ and domain-specific features as $F_d$. The domain-specific features can be formulated as shown in equation 6.

\begin{equation}
F_d = \gamma \cdot \varsigma_{d0}(F_t) + (1 - \gamma) \cdot \varsigma_{d1}(F_t)
\end{equation}
where $\varsigma_{d}$ refers to the branches in the GMU unit and $\gamma$ represents the control factor and comprises of dichotomous values ${0,1}$. The GMU outputs the style code from each of the pre-trained AnimeGANv2 network. The last component of our architecture is the generator network that follows StyleGAN. Formally, the style transfer is achieved by $\mathbb{G}(PEnc_1(I), PEnc_2(I), \mathbb{W})$, where $PEnc_1$ and $PEnc_2$ represent pre-trained encoders from AnimeGANv2 and $\mathbb{W}$ refers to the weight vector having size $\mathbb{R}^{18}$.

The coarse-resolution layers learn to map the high-level shape style transfer while the fine-resolution layers learn the low-level color style, respectively. We follow the training strategy from DualStyleGAN except for a few changes. The first change is that instead of using a single network in an extrinsic style path we use two pre-trained encoders from AnimeGANv2 along with the GMU to extract style codes. The second difference is the use of modulative residual blocks (ModRes) throughout the network instead of using color transform block. The ModRes comprises Residual blocks (ResBlock) \cite{ResBlock} and Adaptive Instance Normalization (AdaIn) \cite{Adain} blocks for simulating changes and style conditioning, respectively.

\subsection{Fine-Tuning Process}
The fine-tuning scheme for style fusion in the proposed work follows the curriculum learning approach \cite{Curriculum}. The fine-tuning process consists of three stages i.e. color, structure, and style transfer. The stage-I strives for color conditioning from one of the pre-trained style networks by model initialization technique, such that the convolutional filters in ModRes from the style code generated by $PEnc_1$ are initialized with zeros, and the filters in ModRes from $PEnc_2$ are initialized with identity matrices. This stage generates a mixed style by transferring colors from the encoded styles while preserving the facial structures from the intrinsic style path. It was observed that the first stage kind of tones up the skin and hair color. It generates more of a bright-toned cartoonish portrait. We believe that the main reason for the said results is the use of AnimeGANv2 with a celeb-distill style, respectively.

The fine-tuning in stage-II fuses mid-level styles that manipulate facial structures in terms of make-up, eyes, and hair color. Random latent codes $\mathfrak{c}_1, \mathfrak{c}_2$ are drawn to approximate the style fusion through $\mathbb{G}(\mathfrak{c}_1, \mathfrak{c}_2, 1)$ with perceptual loss. The fine-tuning process is performed on $g(\mathfrak{c}_\mathfrak{l})$ and random latent codes, while gradually decreasing the layers $\mathfrak{l}$, accordingly. The concatenation of the latent codes with respect to layers is denoted by $\mathfrak{c}_\mathfrak{l}^+$ and $g$ represents the output from $\mathbb{G}$ when $\mathbb{W} = 0$. The objective function for fine-tuning is shown in equation 7.
\begin{equation}
\operatorname*{min}_{\mathbb{G}} \operatorname*{max}_{\mathbb{D}} \mathit{loss}_{pl}(\mathbb{G}(\mathfrak{c}_1, \mathfrak{c}_2, 1),g(\mathfrak{c}_\mathfrak{l}^+)) \cdot \alpha_{pl} + \mathit{loss}_{adv} \cdot \alpha_{adv}
\end{equation}
The objective function will be able to learn more structural styles along with colors by decreasing the layers, accordingly.

\begin{figure}[t]
  \centering
   \includegraphics[width=\linewidth]{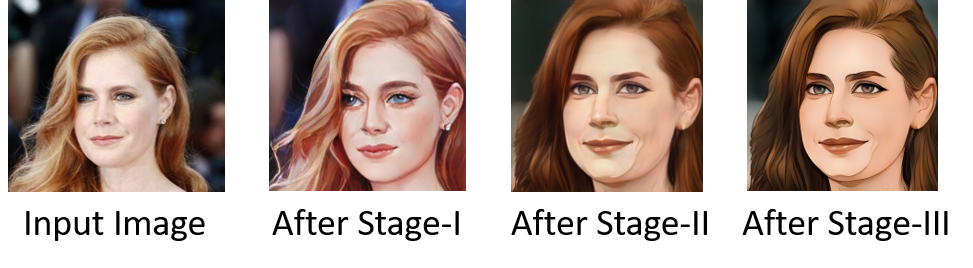}
   \caption{Results of FISTNet after each fine-tuning stage}
   \label{fig:fig3}
\end{figure}

The third stage employs the identity loss and style loss (comprising of feature matching loss \cite{Adain} and contextual loss \cite{Context}) which adds the abstractive styles from $PEnc_2$ and $L_2$ regularization on the ModRes blocks, accordingly. The regularizations help to preserve the facial structure while making the weight values for the residual features close to 0. The full objective function is formulated in equation 8.  
\begin{equation}
\operatorname*{min}_{\mathbb{G}} \operatorname*{max}_{\mathbb{D}} \mathit{loss}_{content} + \mathit{loss}_{style} + \mathit{loss}_{pl} \cdot \alpha_{pl} + \mathit{loss}_{adv} \cdot \alpha_{adv}
\end{equation}
An example of stage-wise generation is shown in Figure \ref{fig:fig3}.

\section{Experiments and Results}
\subsection{Datasets}
One of the motivations for conducting this study is to generate facial style transfer with a limited number of training data. We perform the fusion of the pre-trained AnimeGANv2 for generating styles in face portraits and celeb distill styles. However, we also conducted experiments while considering the combinations of pre-trained Disney and arcane styles to see the effect available with AnimeGANv3. Meanwhile, for the intrinsic style path, we use 317 images from the cartoon dataset \cite{RDGAN}, accordingly. The experiments are conducted on CelebA-HQ \cite{Celeb} dataset to conduct a fair comparison with state-of-the-art works.  

\subsection{Implementation Details for Intrinsic Style Path}
As mentioned, we use the StyleGAN model for fine-tuning the intrinsic style path. Our implementation is based on the PyTorch framework. All the images were resized to 256$\times$256 resolution. The structural loss was applied on the first 2 blocks of the StyleGAN model. We used the same training strategy as \cite{DualStyleGAN}, with an initial learning rate of 0.05. To obtain $\sigma y^*$ we use 10 iterations for the optimization of each sample. The pre-trained face embedding model \cite{FaceRec}, which uses ResNet18 \cite{ResBlock} has been leveraged for $\mathit{loss}_{id}$ and $\mathit{loss}_{seg}$. The value of the regularization parameter $\alpha_{seg}$, is set to 0.2, accordingly. All the aforementioned parameters are selected based on the experiential study while model training process.

\subsection{Implementation Details for Extrinsic Style Path}
The extrinsic style path uses pre-trained encoders from AnimeGANv2 followed by the GMU and generator network. The GMU comprises fully connected domain-specific layers. We use two domain-specific layers in GMU, accordingly. The generator network comprises 18 modulative residuals (ModRes) blocks. There is one 4$\times$4, two 8$\times$8, 16$\times$16, 32$\times$32, 64$\times$64, 128$\times$128, 256$\times$256, 512$\times$512, and 1024$\times$1024 convolution layers and one 1024$\times$1024 to RGB layer in the generator network, respectively. The hyperparameters are the same as DualStyleGAN study. We train the 5th layer for 2k iterations while the 6th and 7th layers for 200 iterations each.

\subsection{Comparison with State-of-the-Art Methods with User Study}
\begin{figure*}[t]
  \centering
\includegraphics[width=\linewidth]{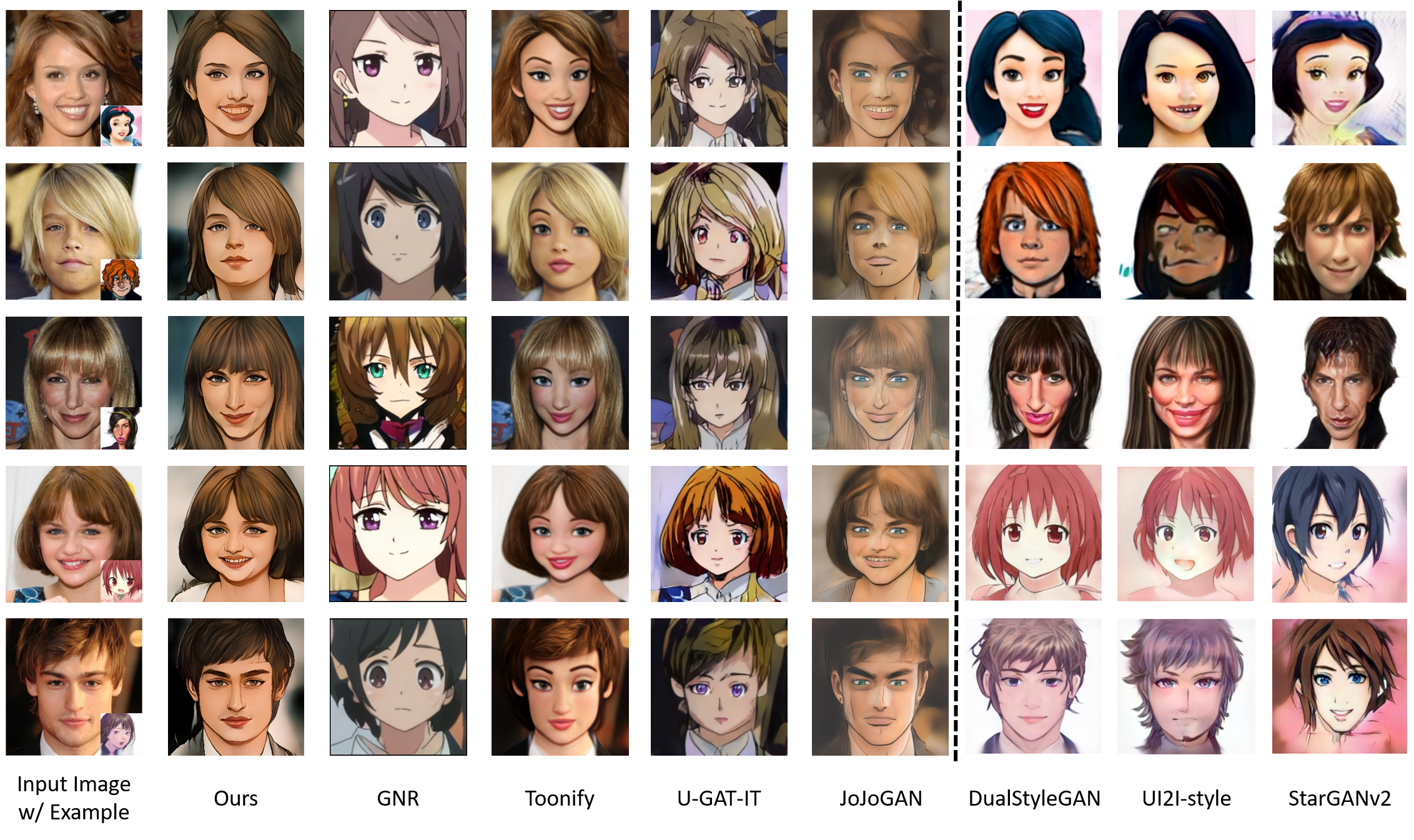}
   \caption{Qualitative Comparison for style transfer with state-of-the-art works. The right side of the line shows the results from existing methods that require an input style while the left side of the line compares with the works that require exemplar images.}
   \label{fig:fig3compare}
\end{figure*}

A qualitative comparison is presented with seven state-of-the-art methods performed in Figure \ref{fig:fig3compare}. Some of them require an example image for style transfer such as DualStyleGAN, UI2I style \cite{StyleGAN2}, and StarGAN v2 \cite{StarGAN}, while other methods either rely on style category or generate a specific style including GNR \cite{GNR}, Toonify \cite{VToonify}, U-GAT-IT \cite{ugatit}, and JoJoGAN \cite{JoJoGAN}, respectively. The results for the aforementioned are generated using their provided codes or APIs (Huggingface or grade io), accordingly. Methods such as U-GAT-IT, Toonify, and GNR do not consider multiple image styles, rather they only rely on domain level to learn the style transfer. JoJoGAN and StarGANv2 ignore the facial structures and details, which results in a kind of overfitting anime or cartoon style. DualStyleGAN and UI2I capture some structural details and color information, but they can need an example image which sometimes generates bad results. Furthermore, none of the results generate a blend of artistic transfer while keeping the facial structure intact. Our proposed work FISTNet generates the best results without providing any example image. However, other styles such as arcane and Disney can be generated by replacing the pre-trained encoders in extrinsic style paths, respectively. It should also be noted the proposed work considers the least amount of data for training.

We also provide a quantitative evaluation by conducting a user study of 67 subjects. Subjects were invited to rate the results generated by the aforementioned methods on three characteristics, which are the preservation of facial details, quality of generated image, and style transfer results. Ten examples were provided for each method during the evaluation. Furthermore, the subjects were allowed to vary three different examples for methods, including DualStyleGAN, UI2I-style, and StarGANv2. Average preference scores from the subjects are reported in Table 1. The proposed method scores best among all the existing methods for facial preservation, image quality, and style quality, respectively. It was also observed that the Toonify approach yields the second-best results.\\

\subsection{Comparison with State-of-the-Art Methods with FID}
The Frechet Inception Distance (FID) \cite{FID} is extensively used in facial style transfer studies to evaluate the diversity and quality of images that are generated through population statistics. Studies such as \cite{JoJoGAN, Ojha} have also used FID for evaluating style mappers. The process for evaluating the facial style transfer quality through FID is as follows:
\begin{itemize}
    \item Select a reference image and perform stylization.
    \item For generalized standard, its better to perform one-shot stylization.
    \item Compute FID between style and the generated result after applying style transfer.
    \item Compute FID using the testset.
\end{itemize}
The process is also compliant with the existing works \cite{JoJoGAN}. For fair comparison, we replaced the pre-trained encoder for celeb-distill style with the sketch one and generated the images, accordingly. The results with sketch generative style using FISTNet are shown in Figure 5, while the FID scores are reported in Table 2, respectively. The lower the FID score, the better facial structures are preserved. The proposed FISTNet achieves the lowest FID score after Ojha et al. \cite{Ojha} in comparison to all the state-of-the-art works. The BlendGAN \cite{BlendGAN} achieves a lower FID than many state-of-the-art works, however, it was noticed that BlendGAN does not perform stylization well on the images. Meanwhile the lowest FID was achieved by Ojha et al. \cite{Ojha} but the results showed that strong distortions were imposed on the faces. The observation concerning facial distortions for the method proposed in Ojha et al. \cite{Ojha} is compliant with the study \cite{JoJoGAN}.

\begin{table}[]
\centering
\caption{User scores for varying characteristics. The best scores are represented with bold. *FP $\rightarrow$ Facial Preservation, IQ $\rightarrow$ Image Quality, SQ $\rightarrow$ Style Quality, Avg $\rightarrow$ Average.}
\label{tab:my-table}
\begin{tabular}{|l|c|c|c|c|}
\hline
Method              & FP            & IQ            & SQ            & Avg            \\ \hline
UI2I-style    & 0.07          & 0.04          & 0.03          & 0.047          \\ \hline
StarGANv2     & 0.09          & 0.07          & 0.03          & 0.063          \\ \hline
GNR           & 0.06          & 0.07          & 0.04          & 0.057          \\ \hline
U-GAT-IT      & 0.06          & 0.08          & 0.03          & 0.057          \\ \hline
DualStyleGAN & 0.13          & 0.15          & 0.18          & 0.153          \\ \hline
JoJoGAN       & 0.17          & 0.05          & 0.04          & 0.087          \\ \hline
\textbf{FISTNet }          & \textbf{0.24} & \textbf{0.33} & \textbf{0.38} & \textbf{0.316} \\ \hline
\end{tabular}
\end{table}

\begin{table}[]
\centering
\caption{Comparative Analysis with state-of-the-art works using FID score. }
\label{tab:my-table}
\begin{tabular}{|c|c|}
\hline
Method       & FID   \\ \hline
U-GAT-IT     & 183.2 \\ \hline
GNR          & 167.4 \\ \hline
BlendGAN     & 94.7  \\ \hline
JoJoGAN      & 107.6 \\ \hline
Ojha et al.  & 74.5  \\ \hline
DualStyleGAN & 155.2 \\ \hline
Toonify      & 79.7  \\ \hline
FISTNet    & 78.9  \\ \hline
\end{tabular}%
\end{table}

\begin{figure*}[t]
  \centering
   \includegraphics[width=\linewidth]{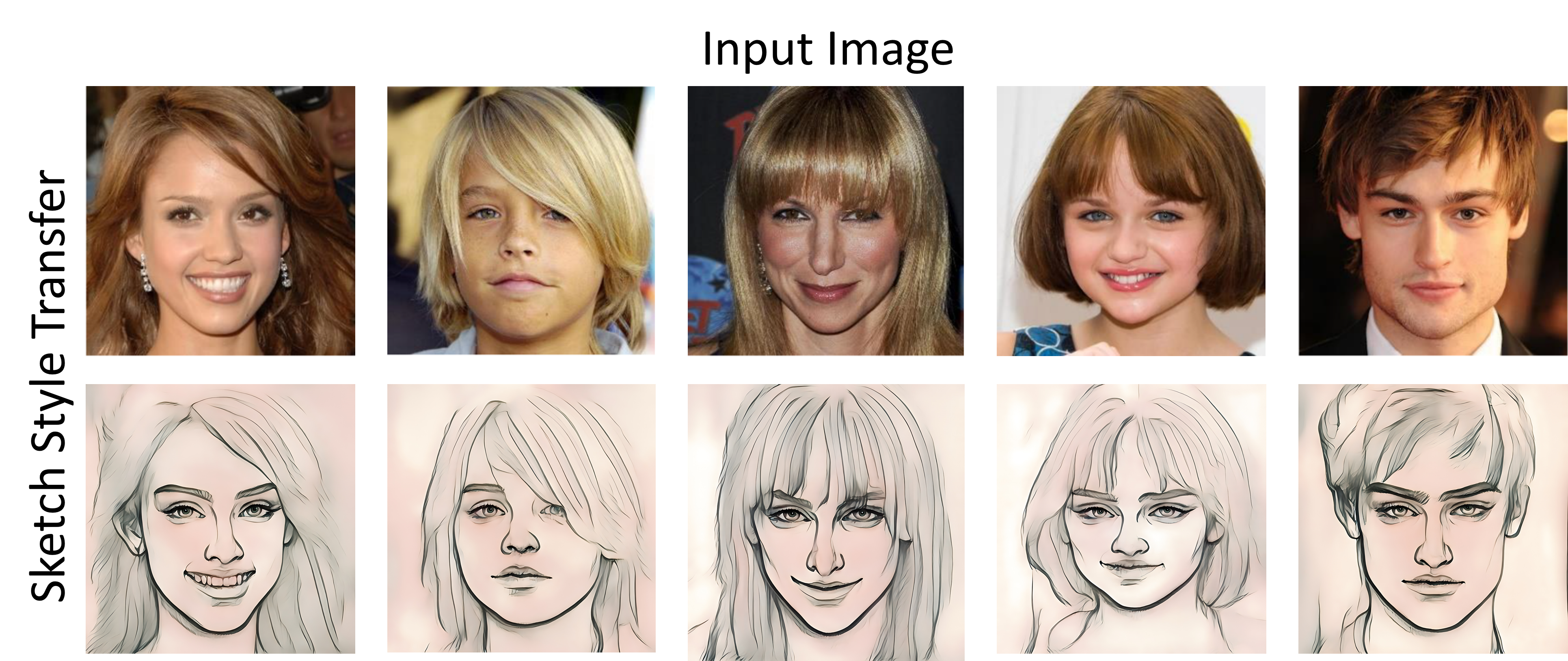}
   \caption{Sketch style transfer using FISTNet}
   \label{fig:fig8}
\end{figure*}

\begin{figure}[t]
  \centering
   \includegraphics[width=\linewidth]{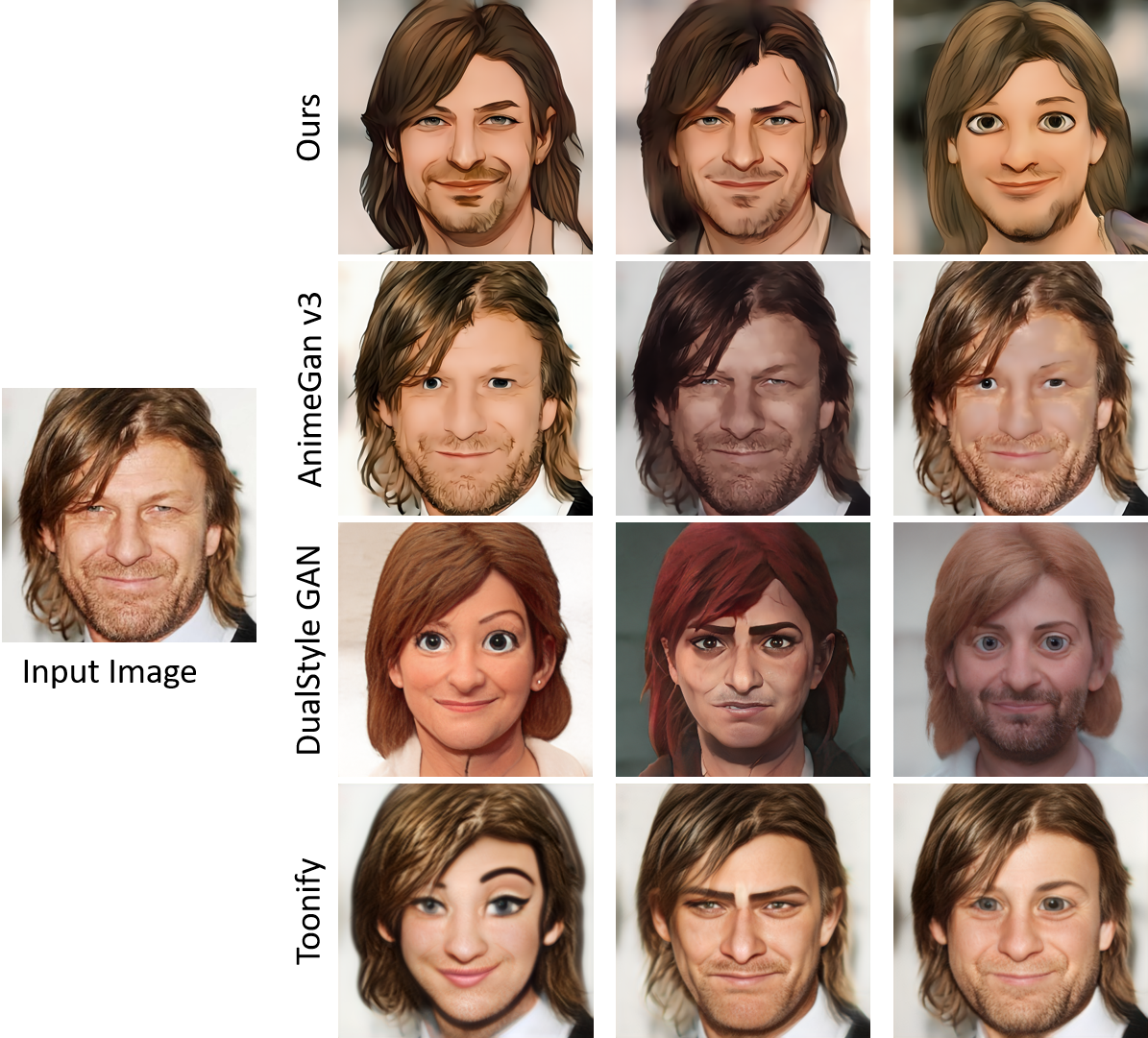}
   \caption{Qualitative comparison on celebrity images with state-of-the-art works using multiple styles}
   \label{fig:fig5}
\end{figure}

\begin{figure}[t]
  \centering
   \includegraphics[width=\linewidth]{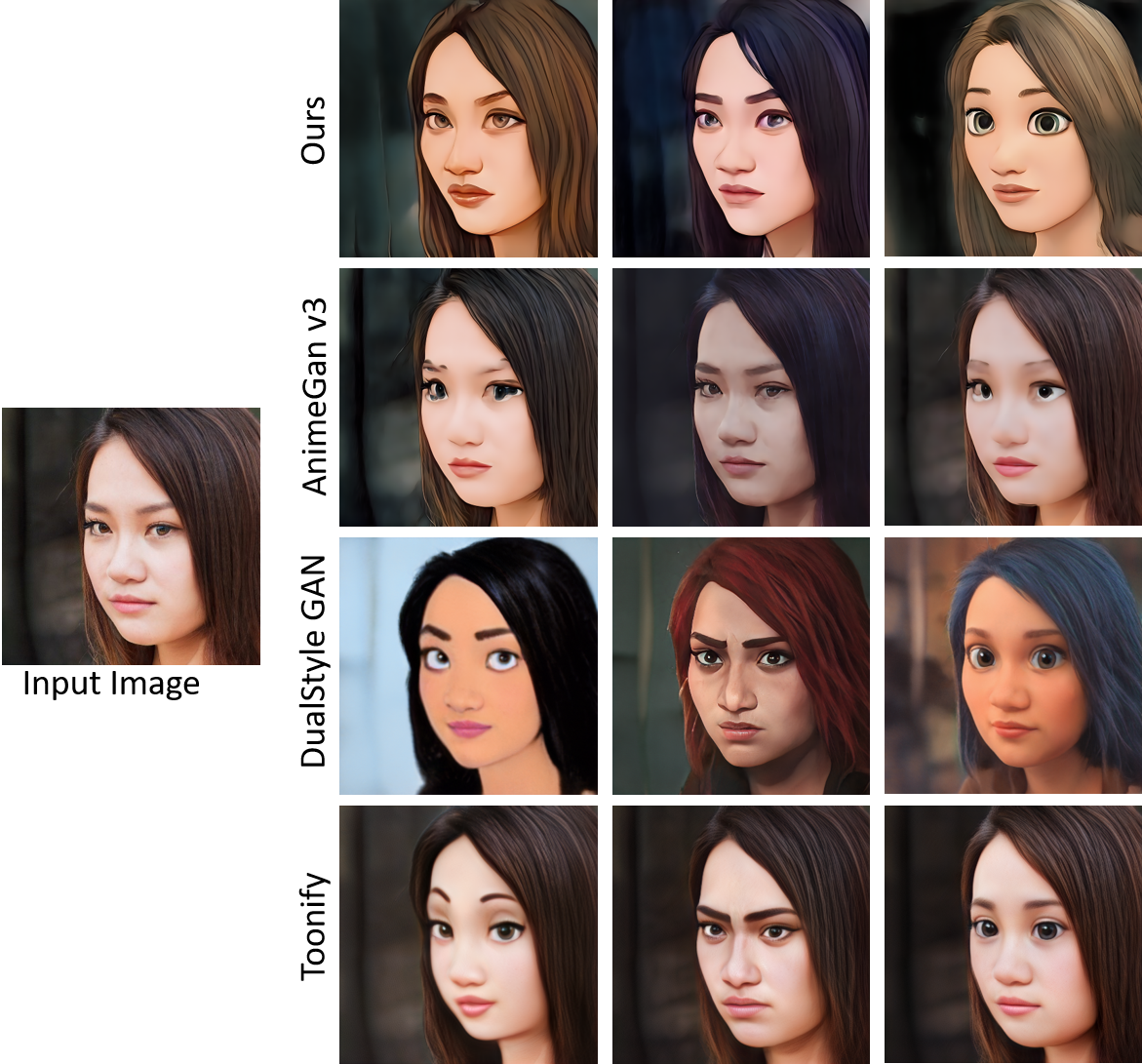}
   \caption{Qualitative comparison on randomly generated images (style-in-the-wild experiment) with state-of-the-art works using multiple styles.}
   \label{fig:fig6}
\end{figure}

\subsection{Comparison with State-of-the-Art on Multiple Styles}
We further compare our approach with state-of-the-art works such as Toonify, AnimeGANv3\footnote{https://github.com/TachibanaYoshino/AnimeGANv3}, and DualStyleGAN for Cartoon, Arcane, and Disney styles. The Toonify and the proposed approach only need to select the style type, while the DualStyleGAN also needs an exemplar image. In this regard, we selected the image index of 142, 40, and 67, for cartoon, arcane, and Pixar (Disney) styles, respectively. The qualitative comparison is shown in Figure \ref{fig:fig5}. The results illustrate the weakness of AnimeGANv3 when tried with Disney style as it adds random artifacts. The Toonify approach retains the facial features better when applying the Disney style in comparison to the proposed work. However, the yielded image quality is on the downside as it's a little bit blurry. In summary, the proposed results yield high-quality images with better visual quality in comparison to the aforementioned existing works across multiple styles. It should also be noted that our work utilizes less data for training in comparison to the existing works.

We repeated the same experiment to test and compare the style transfer results on a randomly generated face\footnote{https://this-person-does-not-exist.com/en}. Similar to the previous experiment, we selected exemplar samples for generating images using DualStyleGAN. The image index for Cartoon, Arcane, and Disney stiles are 183, 10, and 118, respectively. The results are shown in Figure \ref{fig:fig6}. The results highlight the limitation of AnimeGANv3 with randomly generated and sideways faces. Toonify generates good results, but the facial style transfer is limited in comparison to the results on celebrity images. The proposed work not only overcomes the problems associated with AnimeGANv3 but also transfers diverse styles successfully on the randomly generated image.   

\subsection{Limitations}
\begin{figure}[t]
  \centering
   \includegraphics[width=\linewidth]{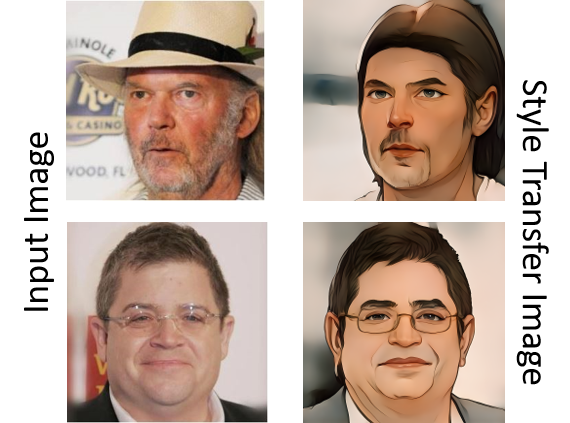}
   \caption{Limitations of FISTNet. The artifacts are introduced when facial props are used in the input images. }
   \label{fig:fig7}
\end{figure}

A few limitations of the work are quite similar to AnimeGANv3 and DualStyleGAN, i.e. the facial features and structure is retained quite well. However, artifacts are introduced due to the props used with images such as caps, hats, glasses, and so forth. We believe that the problem is due to the limited amount of data being used. One such example is shown in Figure \ref{fig:fig7}. One of the problems is also shown in Figure \ref{fig:fig6}, arcane style, where some artifacts are generated under the eyes when the facial image is sideways. The problem is quite consistent with AnimeGANv3, which we believe is due to the adoption of the pre-trained network. Another limitation that is observed in Figure 5 while performing sketch style transfer is that it does not model the smile (especially the toothy smile) well.

\section{Conlusion and Future Work}
This paper proposes DISTNet that perform style conditioning on facial images while preserving facial characteristics and structure. Inspired by the DualStyleGAN, we use intrinsic and extrinsic style paths with StyleGAN for generating diverse stylized facial images. However, contrasting with DualStyleGAN, we use existing pre-trained networks that not only help in adding unique stylization but also eliminate the problem of large-scale training data. We also replaced the transform coding blocks with modulative residual blocks, which help in generating high-image and style-quality images, respectively. We also show that the FISTNet networks are able to generate multiple styles such as arcane and Disney, by using existing pre-trained networks. We believe that the proposed work can be useful for emerging applications such as eXtended Reality (XR), Metaverse, avatar generation, and art generation for non-fungible tokens (NFTs). In the future, we would like to explore more diverse styles and add an extra path for the style transfer to observe the results. Furthermore, we intend to add a localized-stylization approach that will provide users the freedom to perform fusion of multiple styles in a single image.

\bibliography{egbib}

\end{document}